%% file: main.tex
\documentclass[conference]{IEEEtran}

\usepackage{graphicx}
\usepackage{acronym}
\usepackage{subcaption}
\usepackage{hyperref}
\usepackage{xurl}

\acrodef{SNN}[SNN]{Spiking Neural Network}

\def\BibTeX{{\rm B\kern-.05em{\sc i\kern-.025em b}\kern-.08em
    T\kern-.1667em\lower.7ex\hbox{E}\kern-.125emX}}

\begin{document}

\title{Spike-based building blocks for performing logic operations using Spiking Neural Networks on SpiNNaker}

\author{\IEEEauthorblockN{Alvaro Ayuso-Martinez, Daniel Casanueva-Morato, Juan P. Dominguez-Morales, \\ Angel Jimenez-Fernandez and Gabriel Jimenez-Moreno}
\IEEEauthorblockA{\textit{Robotics and Technology of Computers Lab.} \\
\textit{Universidad de Sevilla}\\
Sevilla, España \\
aayuso@us.es}
}

\maketitle

\input{sections/abstract}


\input{sections/introduction}
\input{sections/technologies}
\input{sections/designs}

\input{sections/results}
\input{sections/conclusions}

\input{sections/acknowledgments}

\bibliographystyle{IEEEtran}
\bibliography{sections/bibliography}

\end{document}

%% file: sections/abstract.tex
\begin{abstract}
One of the most interesting and still growing scientific fields is neuromorphic engineering, which is focused on studying and designing hardware and software with the purpose of mimicking the basic principles of biological nervous systems. Currently, there are many research groups developing practical applications based on neuroscientific knowledge. This work provides researchers with a novel toolkit of building blocks based on Spiking Neural Networks that emulate the behavior of different logic gates. These could be very useful in many spike-based applications, since logic gates are the basis of digital circuits. The designs and models proposed are presented and implemented on a SpiNNaker hardware platform. Different experiments were performed in order to validate the expected behavior, and the obtained results are discussed. The functionality of traditional logic gates and the proposed blocks is studied, and the feasibility of the presented approach is discussed.
\end{abstract}

%% file: sections/introduction.tex
\section{Introduction}
\label{intro}

Over the last two decades, neuroscience has experienced a major growth. This growth, along with a set of main objectives of this field of knowledge, are presented in \cite{markram2013seven}. As a consequence, neuromorphic engineering, which was a concept presented by Carver Mead in \cite{mead1990neuromorphic} that focuses on the study, design and implementation of hardware and software with the aim of mimicking the basic principles of biological nervous systems, has become one of the most promising scientific fields. Numerous works have studied the progress of this research field, such as \cite{soman2016recent}, which presents a graph that collects the research activity focused on neuromorphic engineering between the years 2005 and 2015.

In neuromorphic systems, the information is transmitted using spikes, which are asynchronous electric pulses generated by artificial neurons. This approach is very similar to the biological model, where these spikes are large peaks in the membrane potential of neurons that occur when the membrane potential reaches the threshold potential. Representing information with spikes has powerful advantages, such as lower power consumption and better real-time capability than the traditional methods \cite{NeuromorphNature} \cite{zhu2020comprehensiveReview}.

Processing spikes requires a bio-inspired computational approach, which implies the need to work with special hardware and software. A specific type of biologically-plausible neural networks called \acp{SNN} are commonly used for this purpose. These \acp{SNN} have two basic bio-inspired elements: neurons and synapses, where analogizing with the mathematical theory of graphs, neurons would be nodes and synapses would be edges. 

Some of the most striking applications in neuromorphic engineering include the use of \acp{SNN}, such as speech recognition \cite{dominguez2018deep, wu2020deep}, sensory fusion \cite{corradi2021gyro, schoepe2020live}, motor control \cite{stagsted2020towards, glatz2019adaptive} and bio-inspired locomotion \cite{gutierrez2020neuropod, batres2021biologically}, which demonstrates its importance.

Although \acp{SNN} are the basis for most neuromorphic engineering applications, there are no rules on how to build networks in order to achieve a specific behavior. In fact, the process to perform one of these applications is as follows: firstly, the whole biological mechanism must be understood with neuroscientific knowledge; then, a bio-inspired system is developed based on this biological mechanism. However, although much is known about brain functions, this knowledge is sometimes not advanced enough to determine the biological basis of some of its parts, but its functionality. Therefore, many works cannot be currently developed and are proposed to be developed in the future.

A great solution to this type of problem is to design functional blocks with specific behaviors that provide the system with the missing functionality, completing it. These functional blocks are built from so-called building blocks, which are other functional blocks with basic functionality. In digital circuits, these building blocks are logic gates, which are made of transistors and are massively combined in order to create complex digital systems. In this work we propose the use of \acp{SNN} to make high-level designs using neurons and synapses, in a biological approach. Nevertheless, these neuron models are implemented in hardware using transistors since they are essentially digital circuits.


The main objective of this work is to provide a toolkit of bio-inspired logic gates. As in digital circuits, these logic gates must carry out logical operations. Although these operations are supposed to be based on Boolean logic, they are asynchronous operations and much more punctual than their digital equivalents, since the spikes are required to coincide at a given point in time to perform the desired function. This is the basic principle of low power consumption of these gates. The operating frequency will depend on the speed with which it is possible to generate spikes in each of the neurons that make up the entire block, i.e., in their refractory period. Working with small refractory periods should be sufficient to achieve very high operating frequencies, although they are expected to be on the order of milliseconds since a biological approach is intended.

Since we want these spiking logic gates to be totally deterministic, their operation must be completely independent of the neuron parameters. To achieve this, their functionality has to be based on the behavior of the network as a whole, rather than on individual neurons. Static synapses play a fundamental role here, as they allow modifying and creating new behaviors thanks to their delays and weights. 

As \acp{SNN} need to be used to create these building blocks, there are currently different alternatives at both software and hardware level to implement and simulate them. At software level, some very popular simulators can be found, such as NEST \cite{Gewaltig:NEST} and Brian \cite{goodman2008brian}. At hardware level, some of the most outstanding simulation platforms are SpiNNaker \cite{furber2014spinnaker}, Loihi \cite{davies2018loihi} and TrueNorth \cite{akopyan2015truenorth}. In this work, the SpiNNaker hardware platform is used for running large-scale neural network simulations in real time. 

The developed spiking logic gates, which are deeply explained in this paper, are completely available on a public GitHub repository\footnote{\url{https://github.com/alvayus/sPyBlocks/}}, allowing neuromorphic engineers to use them in order to achieve the required functionalities. These new blocks have been deeply tested to ensure that their functionalities are as expected. This work paves the way for higher-level spiking block design in order to achieve more complex specific functionalities.

The rest of the paper is structured as follows: in Section~\ref{technologies}, the software and hardware materials used is detailed in depth; in Section~\ref{designs}, a list of proposed building blocks and other additional functional blocks with their corresponding designs is presented; in Section~\ref{results}, the results obtained from the tests of some of the proposed functional blocks are discussed; and finally, in Section~\ref{conclusions}, the conclusions obtained from the results of the experiments are presented.

%% file: sections/technologies.tex


\section{Material and methods\label{technologies}} 

\subsection{Spiking Neural Networks}

As was mentioned in the introduction, \acp{SNN} are used in order to design functional blocks. These \acp{SNN} are considered the third generation of neural networks and consist in bio-inspired neuron models and synapses. Thus, this is the closest approach of neural networks to biological functioning \cite{davidson2021comparison}. 

Information is transmitted across synapses in the form of spikes, which are asynchronous electric pulses (large peaks in the membrane potential of neurons that occur when the membrane potential reaches the threshold potential) produced by neurons. These output spikes reach other neurons of the neural network as input spikes.

This biological approach of neural networks has powerful advantages, such as low power consumption and a great real-time capability.

\subsection{SpiNNaker}

SpiNNaker is a massively-parallel multi-core computing system which was designed to allow modelling very large \acp{SNN} in real time and whose interconnect architecture is inspired by the connectivity characteristics of the mammalian brain \cite{furber2014spinnaker}. In this work, we use a SpiNN-3 machine to have a limitation in the number of resources used and the obligation to optimize the implemented designs, while also having a SpiNN-5 machine available to test more complex designs. 

In \cite{rowley2019spinntools}, the most important features of the SpiNN-3 and the SpiNN-5 machines are detailed. The SpiNN-3 machine has 4 chips, whereas the SpiNN-5 machine has 48 chips. The SpiNN-5 machine additionally has 3 FPGAs, allowing the machine to be connected to 6 other boards to make up a larger SpiNNaker machine. Each of the chips has 18 ARM968E-S processors/cores, operating at 200 MHz. A 100 Mbps Ethernet connection is used as an I/O interface and to send scripts and commands to the board.

\subsection{PyNN}

In order to develop the proposed public package of spiking logic gates, PyNN was used, which is a Python package for the simulator-independent specification of neuronal network models \cite{davison2009pynn}. Currently, PyNN supports NEURON \cite{hines1997neuron}, NEST \cite{Gewaltig:NEST} and Brian \cite{goodman2008brian} as software neural network simulators, and the SpiNNaker \cite{furber2014spinnaker} and BrainScaleS neuromorphic hardware systems.

\subsection{Neuron parameters}

Although we want the behavior of the networks to be independent of the neuron parameters, two details must be taken into account. Firstly, it is necessary to ensure that neurons will fire at most once when receiving the expected input spikes to achieve the desired behavior of the functional blocks. Secondly, it is necessary to ensure that, after firing, these neurons will not receive any more input spikes until they reach the resting potential. 

Different sets of neuron parameters were used to test the response of a single neuron. This way, the best set of parameters for the designed \acp{SNN} was found, which is the following: $c_m$: 0.1 nF, $tau_m$: 0.1 ms, $tau_{refrac}$: 1.0 ms, $tau_{syn\_E}$: 0.1 ms, $tau_{syn\_I}$: 0.1 ms, $v_{rest}$: -65.0 mV, $v_{reset}$: -65.0 mV, $v_{thresh}$: -64.91 mV, where $c_{m}$ is the membrane capacitance, $tau_{m}$ is the time-constant of the RC circuit, $tau_{refrac}$ is the refractory period,  $tau_{syn\_E}$ and $tau_{syn\_I}$ are the excitatory and inhibitory input current decay time-constant, and $v_{rest}$, $v_{reset}$ and $v_{thresh}$ are the resting, reset and threshold membrane potentials, respectively.

To achieve a high operating frequency, and as was explained in this subsection, it is necessary to quickly reach the resting potential after firing a spike. The most important parameters for this purpose are three of the four time-constants: $tau_m$, $tau_{syn\_E}$ and $tau_{syn\_I}$. The lower these values are, the faster the resting potential is reached. The last of the time-constants, $tau_{refrac}$, is responsible for allowing or avoiding firing when the threshold potential is reached, thus, varying the maximum firing frequency of the neuron. To adjust the potential at which the spike is fired, the membrane potential parameters, $v_{rest}$, $v_{reset}$ and $v_{thresh}$, as well as the membrane capacitance, $c_m$, can be used.

%% file: sections/designs.tex
\section{Designs\label{designs}}
This section shows the architecture of the proposed building blocks, which can be used to create more complex functional blocks. The list of these building blocks is ordered from least to greatest complexity and dependency. Thus, the simplest block will be the first in the list, which is as follows: OR, AND (Classic), SR Latch, Switch (Asynchronous oscillator), XOR, Constant spike source, NOT, Synchronous oscillator, AND (Fast), Flank detector.

\begin{figure*}[htbp]
\centerline{\includegraphics[width=\linewidth]{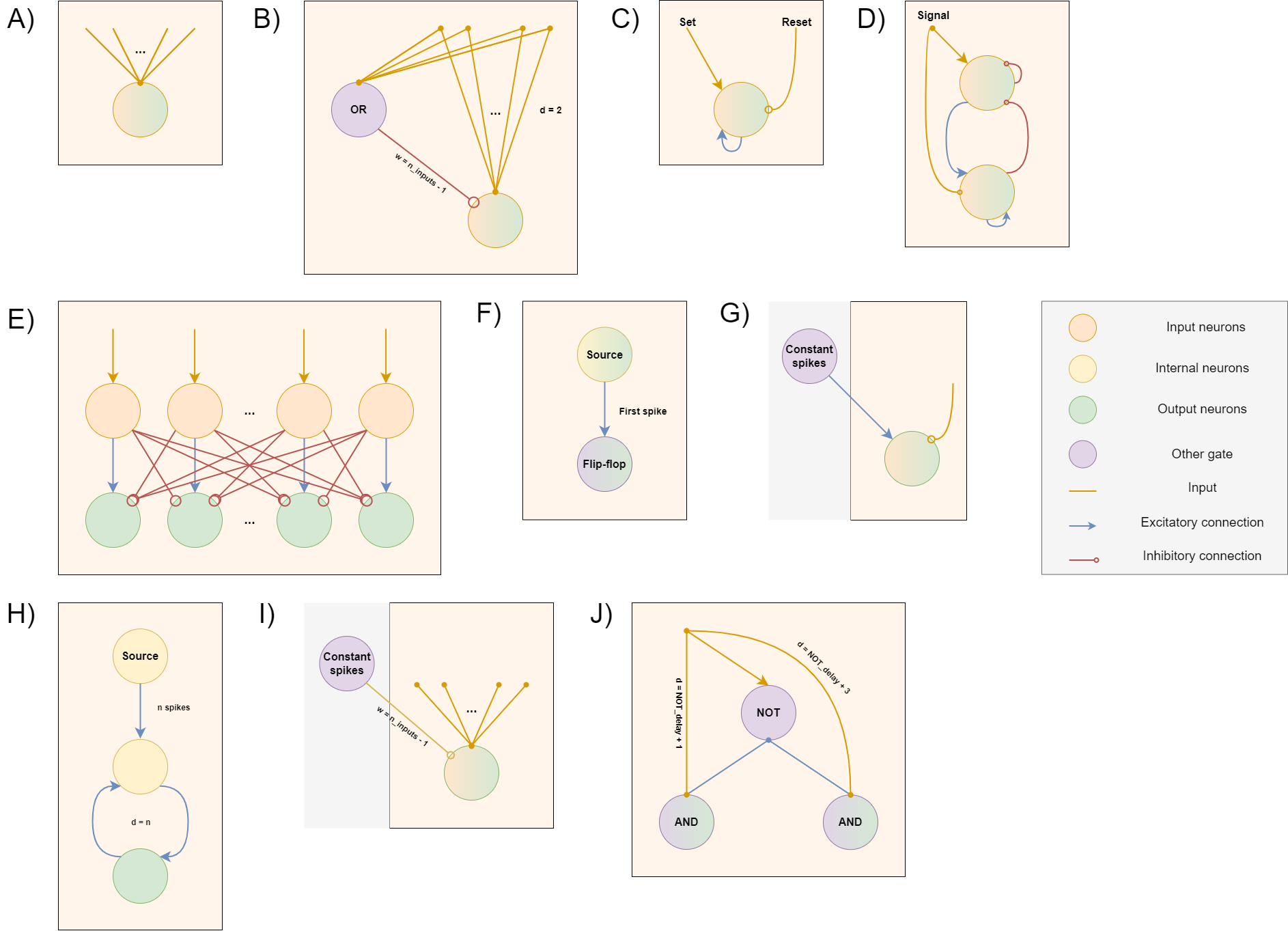}}
\caption{Diagram showing the design of each implemented block. The legend on the right shows what each color and symbol means. \textbf{A)} OR. \textbf{B)} Classic AND. \textbf{C)} SR Latch. \textbf{D)} Switch. \textbf{E)} XOR. \textbf{F)} Constant spike source. \textbf{G)} NOT. \textbf{H)} Synchronous oscillator. \textbf{I)} Fast AND. \textbf{J)} Flank detector.}
\label{components}
\end{figure*}

The design of each implemented block can be found in Fig.~\ref{components}, together with a legend of the meaning of each color and symbol used. Input neurons, colored in orange, are the target of input connections, which generally come from the output neurons of other functional blocks.

\subsection{OR}

The OR block (Fig.~\ref{components}A) consists of a single neuron that receives as many inputs as needed. Note that, as the number of input spikes increases, the voltage of the output neuron also increases, meaning that the neuron could fire more than once. It is possible to prevent this behavior by decreasing the time it takes to reach the resting potential, which can be done as explained in Section~\ref{technologies}. This should lower the membrane potential fast enough so that it does not fire again before the end of the refractory period.

\subsection{AND (Classic)}

In this work, this is the first implementation of a block that simulates the AND gate behavior. It consists of an OR block that fires when it receives at least one input spike, producing an inhibition with a weight of $n - 1$ (where $n$ is the number of inputs) in the output neuron, colored in green in Fig.~\ref{components}B. The output neuron receives the same inputs as the OR neuron (delayed to match in time with the output of the OR gate).

Note that the output neuron will fire only when the number of its input spikes is greater than the inhibition produced by the OR neuron, i.e., when it receives $n$ input spikes at the same time.

\subsection{SR Latch}

The bistable circuit is the simplest memory unit. It can be implemented with a single neuron holding or releasing a spike (as can be seen in Fig.~\ref{components}C). A self-excitatory connection is used to keep the input spike received through the set connection cycling around the neuron. This cycling spike is released when the neuron receives an input spike through the inhibitory reset connection. 

Since the circuit is conceived as an asynchronous circuit, this bistable circuit is called a latch. Having a set input and a reset input, this block is called SR latch.

One problem to consider is that the spike cyclically contained in the neuron seems to be violating the law of conservation of energy, although there are biological evidences of the existence of recurrent neural networks in the human brain which use this same principle \cite{douglas2007recurrent}. This is not an obvious problem and would require a much deeper study of neurons and spikes at the biological level.

\subsection{Switch}

This block (Fig.~\ref{components}D) does not include set or reset connections, but a single input signal through which a spike causes the switch to change its state depending on whether or not a spike is being held.

If the inhibitory connections did not exist, the resulting block would be very similar to the SR latch block. In fact, the excitatory connections provide the switch with the same behavior: the input spike causes the upper neuron to fire, and the resulting spike activates the neuron below, which essentially is the SR latch.

Thanks to the inhibitory connections, each spike produces an inhibition in the upper neuron, and thus, the next input spike will be suppressed at that neuron. At the same time, the input spike produces and inhibition in the lower neuron, which releases its held spike, causing the state to change. Making an analogy with digital circuits, the logical one and logical zero would mean whether or not there is a held spike, respectively. 

Note that both neurons are half colored in green, which means that both are output neurons. If the neuron below was the only output neuron, the first output spike of the set state of the switch would be missing.

This switch block is sometimes referred to as an asynchronous oscillator.

\subsection{XOR}

This implementation (Fig.~\ref{components}E) has as many input neurons as input connections, which excite their associated output neuron and inhibit the rest of output neurons when they fire. Thanks to this architecture, there can only be one output neuron firing at the same time, since an output neuron is excited only by its associated input neuron and can be inhibited by the remaining $n - 1$ input neurons. Note that if two or more input neurons fire at the same time, output neurons will not fire.

This is one of the most expensive designs, since it needs a large amount of connections between input and output neurons. More specifically, the number of connections is $n^2 + n$, where $n$ is the number of input connections.

\subsection{Constant spike source}

Despite the counterintuitive idea of having a constant source of spikes, this is needed to implement some functional blocks, especially one of them: the NOT gate. 

As \acp{SNN} are being used, the only way to make a neuron fire is providing it with spikes. There is no way to make a block that generates spikes if it is not receiving them, which is the behavior of the NOT gate. Therefore, we need to implement a block that constantly provides spikes, although, in practice, it could be replaced by a set of connections with other neurons that ensure this constant rate of spikes.

This implementation (Fig.~\ref{components}F) uses a spike source, which is a spike generator and is frequently used in the testing of \acp{SNN}. In this case, it generates only one spike at the beginning of the simulation, which is used to set the SR latch and is also used as the first output spike of the block. Thus, the output of this block will be the output of the SR latch, except for the first output spike, which will be generated by the source.

\subsection{NOT}

Once there is a constant source of input spikes, it is easy to implement a block with the NOT gate behavior. In this implementation (Fig.~\ref{components}G), this constant spike source excites the NOT output neuron, while the input connection provides it with inhibition. Thus, the output will be a constant rate of spikes if there is no input spike, and the output neuron will not fire if there is an input spike.

Note that the constant spike source is not part of the NOT block, since it has been externally defined.

\subsection{Synchronous oscillator}

This block (Fig.~\ref{components}H) is closely related to the switch block, as both could be considered oscillators. The difference between the two is that, in the asynchronous oscillator (switch) block, it is possible to decide when to change the value, which is not possible in the synchronous model.

Since the information is being coded in spikes, this block does not seem very useful, and it should be preferable to work with the asynchronous model, although it could be considered as a good tool for making analogies between digital circuits and these "circuits" based on \ac{SNN} blocks. Despite this, we will use this synchronous model to generate a spike train with a specific length, which will be constantly appearing and disappearing, as a clock signal, and which will be useful to test the flank detector.

The number of spikes contained in the spike train multiplied by the interspike interval will be the half-period of the "signal". To generate these spikes, a spike source object was used. This spike train will be the input of a neuron that will be cyclically connected to another one by two delayed connections. Their delays will be the value of the half-period of the "signal", causing the output neuron to fire only during one half-period, not firing in the second half-period.

\subsection{AND (Fast)}

Despite the problems previously mentioned, it is possible to use the constant spike source block in order to improve some of the other implemented blocks. 

One of the most interesting improvements would be on the AND block (classic implementation), since it is expected to be widely used for building more complex models. Using the constant spike source block allows removing the OR block and its input connections, thus also removing their delays. As the OR block is removed, it is not necessary to make use of delayed connections. This is why this implementation (Fig.~\ref{components}I) is called "fast".

\subsection{Flank detector}

This block (Fig.~\ref{components}J) combines two of the ones that were previously defined: NOT and AND blocks. It mainly allows detecting rising and falling edges through an AND operation with two spikes, combined with the NOT operation. Let the value one be the existence of spike and the value zero the opposite case, a rising edge is considered as the change from zero to one, while a falling edge is considered as the change from one to zero, in a short time interval. This interval will be assumed to be equal to 1 ms at this time, which is also the delay of the synapses.

In the case of rising edges (left AND), the previous value (the value zero) will pass through the NOT gate and will arrive at the AND gate after 2 ms, since the delay of the NOT gate is 0 ms. As the next value arrives 1 ms after the value zero, the left connection will not need to be delayed. Thus, the value one (the spike) will arrive at the same time as the NOT output.

In the case of falling edges (right AND) it is very similar, but the right connection will be delayed 1 ms more than the total delay of the NOT path, as the next value is being used. As the delay of the NOT gate is 0 ms, an input spike will arrive at the right AND gate through the NOT path 2 ms after being fired. Thus, the next value synapse will be delayed 3 ms in order to coincide in time with the NOT output.

This flank detector block is not only useful to test the synchronous oscillator, but it could be used to detect whether there are two consecutive spikes or not, and whether the spike has appeared (zero to one) or disappeared (one to zero) in any block or connection.

%% file: sections/results.tex
\section{Results\label{results}}

Different experiments were performed in order to prove the correct operation of the proposed functional blocks. In this section, some of the most interesting ones are shown. Since there is a hierarchical dependency between these blocks, we will focus on the most complex of them: AND, XOR, switch and the flank detector blocks. In addition, the results obtained with the classic and fast AND implementations are compared in subsection \ref{and_results}. Finally, the resources needed to implement the presented blocks are shown and discussed.

\subsection{AND\label{and_results}}

One way to test the correct operation of these functional blocks is creating arrays of spikes that will be fired at discrete times and checking the associated output responses. There will be as many arrays as input connections, thus each array contains spikes that will be fired through each input connection. 

In the case of AND blocks, only when the number of input spikes received by the output neuron at a specific time is greater than the number of input connections minus one, which is the inhibition generated by the OR neuron (classic implementation) or the constant spike source block (fast implementation), the output neuron will fire. 

\begin{figure}[htbp]
\centerline{\includegraphics[width=\linewidth]{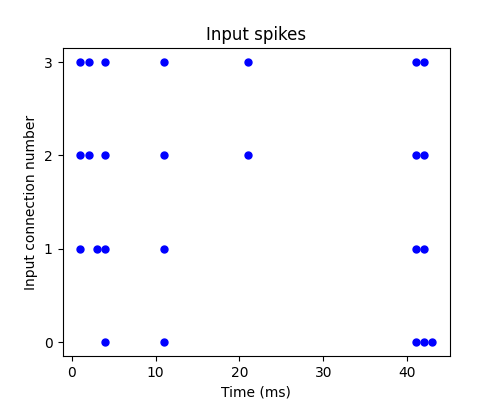}}
\caption{Plot showing the input spikes for each input connection used for testing the AND block.}
\label{and_inputs}
\end{figure}

In this first test, four input connections were used. Fig.~\ref{and_inputs} shows the input spikes for each of them.

\begin{figure}[htbp]
\centerline{\includegraphics[width=\linewidth]{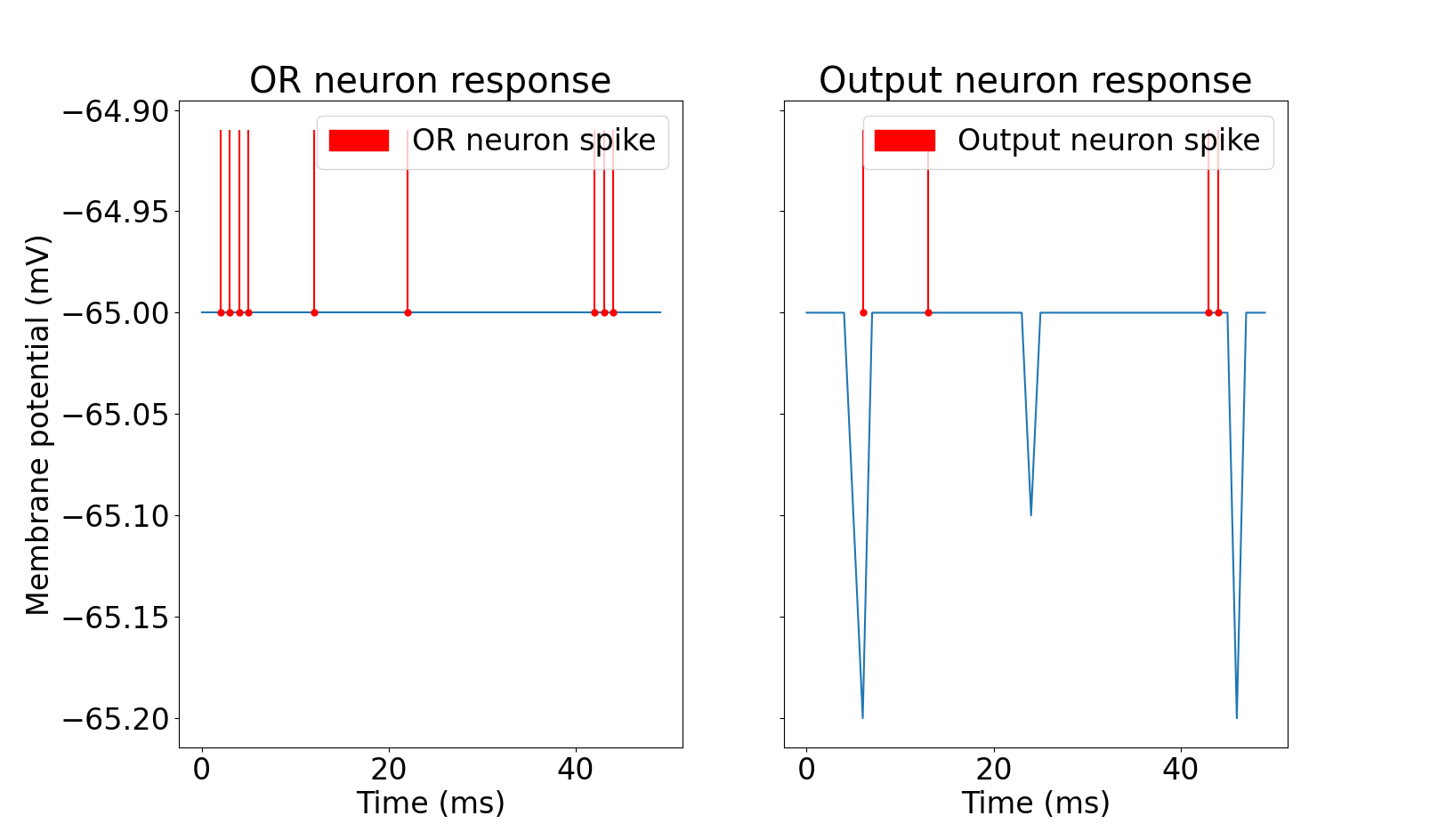}}
\caption{Graphs showing the potential responses and the spikes fired by the OR and output neurons of the AND block (classic implementation). The negative peaks in the output neuron response graph correspond to times when inhibition is greater than excitation.}
\label{and_classic_graph}
\end{figure}

\begin{figure}[htbp]
\centerline{\includegraphics[width=\linewidth]{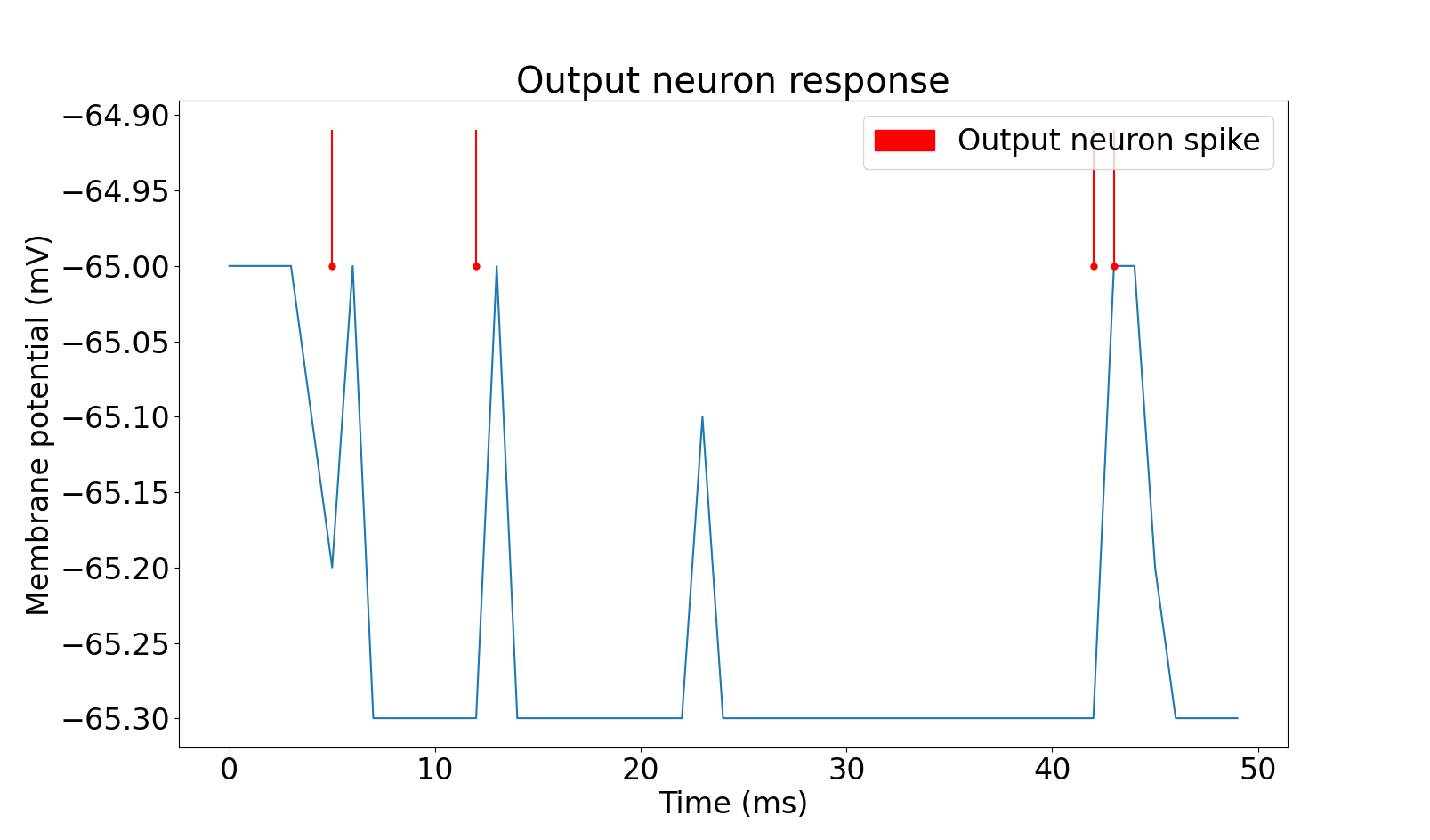}}
\caption{Plot showing the potential responses and the spikes fired by the output neuron of the AND block (fast implementation). In this case, there are positive peaks, since the fast implementation make use of a constant inhibition and occasional excitation.}
\label{and_fast_graph}
\end{figure}

As the AND block is intended to be one of the most widely used blocks, it is interesting to analyze the potential responses of the neurons involved, which, in the case of the classic implementation, are the OR neuron and the output neuron, and, only the output neuron in the case of the fast implementation. Fig.~\ref{and_classic_graph} shows the potential responses and the spikes fired by the OR and output neurons in the classic implementation, while Fig.~\ref{and_fast_graph} shows the potential response and the spikes fired by the output neuron in the fast implementation.

\begin{figure}[htbp]
\centerline{\includegraphics[width=\linewidth]{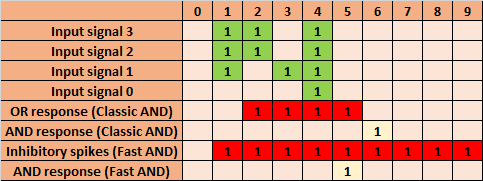}}
\caption{Trace of the AND block test that shows the spikes fired by each neuron of the classic and fast implementations.}
\label{and_trace}
\end{figure}

To have greater control over the results obtained, we automated the generation of spike traces for each test. Fig.~\ref{and_trace} shows the trace obtained in the time range from 0 to 9 milliseconds. In this trace, input signals refers to the spikes of each input connection. Each row of the trace shows a "1" only when there is a spike fired at the millisecond indicated by the top row.

As is explained in Section~\ref{designs}, and as shown in the test trace, the output neuron of the AND block only fires when the number of input spikes is $n$, which in this case is four. Otherwise, no output spike is generated. 

When the number of input spikes is equal to $n - 1$, the membrane potential of the output neuron is equal to the resting potential, since the inhibition is as strong as the excitation for both implementations. 

In the classic implementation, when the number of inputs is less than $n - 1$, the membrane potential of the output neuron reaches negative peaks from the resting potential, as the inhibition is greater than the excitation. On the other hand, in the fast implementation, the membrane potential of the output neuron reaches positive peaks from an "inhibited resting potential", since the output neuron is constantly inhibited and punctually excited. These positive peaks will not be higher than the resting potential in this case, but equal in the case of $n - 1$ input spikes or higher in the case of $n$ input spikes.

In the fast implementation, note that the red spikes shown always coincide with an increasing potential. This potential is not accurate since we work with a "high" firing rate and the time resolution of SpiNNaker is limited to 1 ms. Therefore, from now on we will be only working with spike traces.

Finally, note that in Fig.~\ref{and_trace}, the OR response of the classic implementation is delayed 1 ms from the spike time, while the output spike is delayed 2 ms. The latter delay is the delay of the connection between input spikes and the output neuron, which is 2 ms to match in time with the inhibition generated by the OR neuron. As the time resolution cannot be decreased from 1 ms, this is the lowest possible delay value.

The fast implementation reduces the number of connections as much as possible, only using one set of connections (input connections of the output neuron). As a result, it can be seen that the output spike is delayed only 1 ms from the spike time, which is the delay of these connections.

\subsection{XOR}

\begin{figure}[htbp]
\centerline{\includegraphics[width=\linewidth]{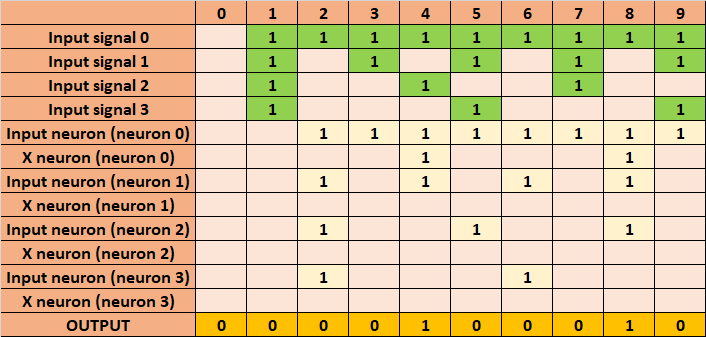}}
\caption{Trace of the XOR block test that shows the spikes fired by each input and output neuron.}
\label{xor_trace}
\end{figure}

To prove the correct operation of the presented XOR block, the proposed test consists in using four signals with different periodicity, i.e., connections through which spikes pass at different times, as input signals for an XOR block with four inputs. The resulting trace is shown in Fig.~\ref{xor_trace}.

In this figure, the periodicity of all input signals is a multiple of the periodicity of input signal 0, which is constantly firing spikes. As the XOR block outputs a new spike only when one input signal is firing at the same time, this occurs when only the input signal 0 is firing. Notice that there is a delay between values in input signal 0 and the output values in the trace, which is related to the delay of the XOR block. The output spikes of the block are the union of the output spikes of each output neuron, which appears as X neuron. 

\subsection{Switch}

\begin{figure}[htbp]
\centerline{\includegraphics[width=\linewidth]{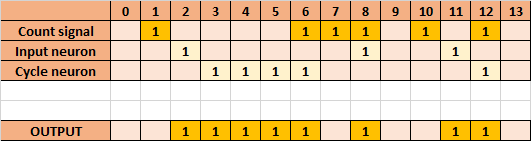}}
\caption{Trace of the switch block test showing the spikes fired by each neuron.}
\label{switch_trace}
\end{figure}

In the same way as with the AND block, the obtained trace for the switch block is shown in Fig.~\ref{switch_trace}.

As is explained in Section~\ref{designs}, the first input spike excites the input neuron, making it fire and propagating the spike to the cycle neuron, which holds it cyclically. Note that it takes 1 ms for the input neuron to fire and 1 ms more for the cycle neuron to start firing due to the delay of the connection. 

The spikes at milliseconds 6, 7 and 8 are used in order to test how quickly the implemented switch is able to change its state. At time 7 ms, the first of these three spikes inhibits the cycle neuron, which releases the held spike. At time 8 ms, the set up mechanism begins again. Finally, at time 9 ms, the input neuron is inhibited. The cycle neuron has just received the output spike from the input neuron, thus it cannot generate inhibition over the input neuron. This justifies the existence of a self-inhibitory connection in the input neuron.

\subsection{Flank detector}

\begin{figure*}[htbp]
\centerline{\includegraphics[width=\linewidth]{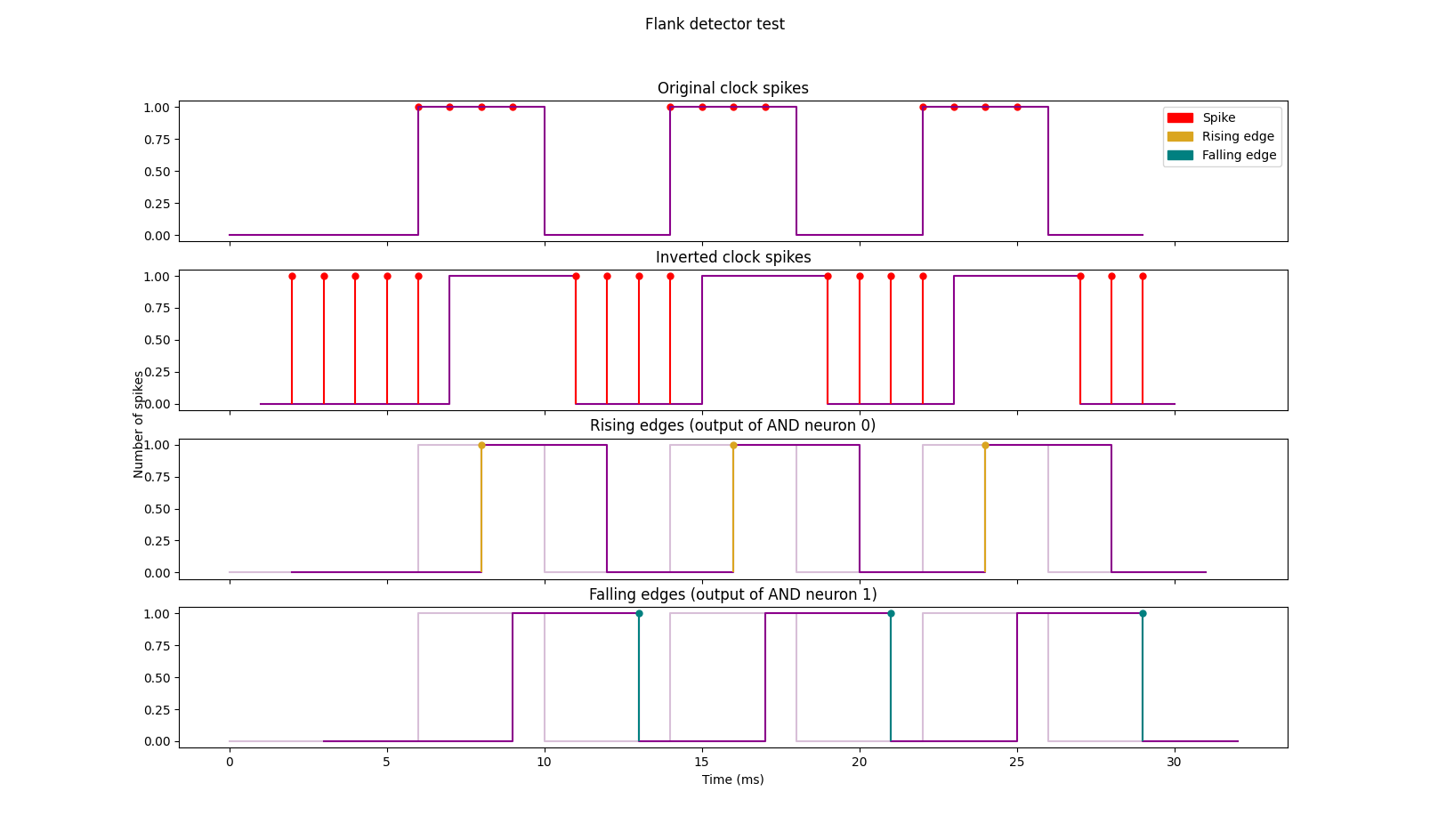}}
\caption{Plot showing the results of the flank detector test.}
\label{flank_detector_graph}
\end{figure*}

As is mentioned in Section~\ref{designs}, to prove the correct operation of the flank detector block, a synchronous oscillator, which is a clock generator, is used.  Fig.~\ref{flank_detector_graph} presents the results of one of the experiments performed to test the functionality of this module. 

In this figure, the top subgraph shows the input spikes, which is a spike train that is continuously oscillating. The subgraph below shows the inverted response of the input spikes, which is the behavior of the NOT block inside the flank detector block. In the lower part of the graph, the detected rising and falling edges are shown, which are the output spikes of the left AND block (rising edge, in orange) and the right AND block (falling edge, in teal blue). The original "clock signal" is shown in each subgraph with the corresponding delay in opaque dark magenta and without delay in translucent dark magenta. 

Note that NOT responses are delayed by 1 ms, rising edges are delayed by 2 ms and falling edges are delayed by 3 ms from the input spike times. These delays are, as in other experiments, due to connection delays. 

Rising edges (from zero to one) are detected as the coincidence between the last spike from the spike train of the NOT response and the first spike from the spike train of the "clock signal", as NOT responses are inherently delayed from the spike time due to the connection delay. The spike of the NOT response that coincides with the spike of the "clock signal" represents the previous value, i.e., the value zero. Rising edges are delayed by 2 ms from the input spike times, since the connection between the NOT block and the AND block adds 1 ms delay and the AND block adds 1 ms extra delay. 

Falling edges (from one to zero) should be detected as the coincidence between the last spike from the spike train of the "clock signal" and the first spike from the spike train of the NOT response. To make the two spikes coincide at the AND gate, the first spike has to be delayed 3 ms (2 ms plus 1 ms of connection delay between the NOT block and the AND block). Since the AND block adds 1 ms extra delay, the time interval between the first spike and the AND output is 4 ms, but 3 ms from the original falling edge.

\begin{figure}[htbp]
\centerline{\includegraphics[width=\linewidth]{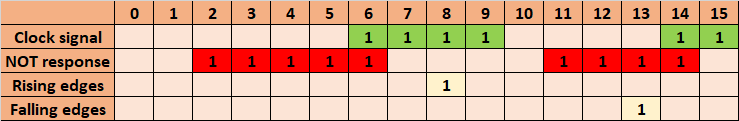}}
\caption{Extract of the trace of the flank detector block test showing the spikes fired by each neuron.}
\label{flank_detector_trace}
\end{figure}

Fig.~\ref{flank_detector_trace} shows an extract of the trace of the flank detector test where both rising and falling edges can be seen. 

This figure shows how the NOT block begins to fire at 2 ms, since its spike source fires at 1 ms. As there is a delay of 1 ms between the synchronous oscillator and the NOT block, at 6 ms the "clock signal" is set to one and this is reflected as a value zero in the NOT block at millisecond 7. Since the AND adds 1 ms extra delay, the rising edge is detected at millisecond 8 (it should be detected at 7, 1 ms after the spike of the "clock signal" and the NOT block matches).

At millisecond 10, the "clock signal" changes from one to zero. This is reflected at millisecond 11 (1 ms later) in the NOT block. As the last spike of the "clock signal" (shown in red) is delayed 3 ms, the falling edge should be detected at millisecond 12. Due to the extra delay of the AND, the spike is detected at millisecond 13.

\subsection{Resources}

Table~\ref{resources_table} shows the resources needed for each block, which include the number of expected inputs, the number of neurons and connections and the latency. $n$ value refers to the number of inputs for one block. Note that the values in the table include the total number of resources, including the resources of the blocks it contains. 

\begin{table}
    \caption{Table showing the resources needed to implement each functional block.}
    \label{resources_table}
    \resizebox{\linewidth}{!}{\begin{tabular}{|c|c|c|c|c|}
    \hline
    \textbf{Block} & \textbf{Expected inputs} & \textbf{Neurons} & \textbf{Connections} & \textbf{Latency (ms)} \\ 
    \hline
    OR & $n$ & $1$ & $n$ & $1$ \\ 
    \hline
    AND (classic) & $n$ & $2$ & $2n + 1$ & $2$ \\ 
    \hline
    SR Latch & $1$ & $1$ & $2 - 3$ & $1$ \\ 
    \hline
    Switch & $1$ & $2$ & $6$ & $1$ \\ 
    \hline
    XOR & $n$ & $2n$ & $n^2 + n$ & $2$ \\ 
    \hline
    \begin{tabular}[c]{@{}c@{}}Constant spike\\ source\end{tabular} & $0$ & $2$ & $2$ & $1$ \\ 
    \hline
    NOT & $1$ & $3$ & $4$ & $1$ \\ 
    \hline
    \begin{tabular}[c]{@{}c@{}}Synchronous\\ oscillator\end{tabular} & $0$ & $3$ & $3$ & $1*$ \\ 
    \hline
    AND (fast) & $n$ & $3$ & $n + 3$ & $1$ \\ 
    \hline
    Flank detector & $1$ & $5$ & $14$ & $2/3$ \\ 
    \hline
    \end{tabular}}
\end{table}

It is important to optimize the number of neurons and synapses on these models, since each block can be used massively in the construction of more complex blocks. Note that the number of connections in the SR latch depends on the existence of set and reset connections. On the other hand, as was previously mentioned, the XOR block is the most expensive by far.

Constant spike sources and synchronous oscillators do not expect any input. In addition, it is shown that synchronous oscillators have a latency of 1 ms, which does not coincide with the delays shown in Fig.~\ref{components}. This is due to the fact that the output is expected to be zero during the first half-period.

In the flank detector (Fig.~\ref{components}J), rising edges (left AND) are delayed 2 ms while falling edges (right AND) are delayed 3~ms.


%% file: sections/conclusions.tex
\section{Conclusions\label{conclusions}}

In this paper, we proposed a list of different functional blocks implemented using \acp{SNN} on SpiNNaker, which can be used to implement desired functionalities.

Digital logic gates and the proposed building blocks are very similar at a high level, since they have the same function. Nevertheless, as the latter are built from \acp{SNN}, they can be considered bio-inspired, including those with recurrent connections which have an analogous biological counterpart \cite{douglas2007recurrent}. Thanks to information being coded using spikes, these spiking logic gates have some advantages over digital logic gates, such as lower power consumption and better real-time capability. Generally, the simplicity of the functional blocks directly affects their real-time capability. Thus, a simple block is considered to be one that has few neurons and connections, and also avoids making use of delayed connections. The smaller the total delay, the better for being used in real time.

All the implemented modules were tested in a set of experiments, demonstrating the expected behavior for each of them and analyzing their requirements in terms of resources needed. The number of neurons and connections of the implemented blocks were optimized, since these modules could be massively used to build higher-level blocks with a more complex functionality.



Future works will focus on using the presented blocks to build more complex functional blocks with higher-level functionalities. Thus, building registers from latches, as well as other components such as decoders or multiplexers using these building blocks, would be straightforward. These new functional blocks could be very useful to implement other higher-level blocks, such as memories, or even a spiking computer which could be the spiking equivalent to the traditional digital computer. In this regard, all the functionalities of a digital computer could be performed with these new architectures, benefiting from the low-power and low-latency nature of spiking neural networks.

The code for the designs and tests of the proposed blocks is available on GitHub. The link to the repository has been attached in the Section~\ref{intro}.

%% file: sections/acknowledgments.tex
\section*{Acknowledgments}
This research was partially supported by the Spanish grant MINDROB (PID2019-105556GB-C33/AEI/10.13039/501100011033). Daniel Casanueva-Morato was supported by a "Formación de Profesorado Universitario" Scholarship from the Spanish Ministry of Education, Culture and Sport.